# A Naturalistic Database of Thermal Emotional Facial Expressions and Effects of Induced Emotions on Memory

Anna Esposito[1], Vincenzo Capuano[1], Jiri Mekyska[2], Marcos Faundez-Zanuy[3]

[1] Second University of Naples, Department of Psychology, Caserta, and IIASS, Italy
[2] Brno University of Technology, Faculty of Electrical Engineering and Communication, Department of Telecommunications, Brno, Czech Republic
[3] EUP Mataró, Avda. Puig i Cadafalch 101, 08303 Mataró (Barcelona), Spain
iiass.annaesp@tin.it, vincenzo.capuano85@gmail.com;
xmekys01@stud.feec.vutbr.cz, faundez@eupmt.es,

**Abstract.** This work defines a procedure for collecting naturally induced emotional facial expressions through the vision of movie excerpts with high emotional contents and reports experimental data ascertaining the effects of emotions on memory word recognition tasks. The induced emotional states include the four basic emotions of sadness, disgust, happiness, and surprise, as well as the neutral emotional state. The resulting database contains both thermal and visible emotional facial expressions, portrayed by forty Italian subjects and simultaneously acquired by appropriately synchronizing a thermal and a standard visible camera. Each subject's recording session lasted 45 minutes, allowing for each mode (thermal or visible) to collect a minimum of 2000 facial expressions from which a minimum of 400 were selected as highly expressive of each emotion category. The database is available to the scientific community and can be obtained contacting one of the authors. For this pilot study, it was found that emotions and/or emotion categories do not affect individual performance on memory word recognition tasks and temperature changes in the face or in some regions of it do not discriminate among emotional states.

## 1 Introduction

The testing of competitive algorithms through data shared by dozens of research laboratories is a milestone for getting significant technological advances [9]. Shared databases allow to validate and develop new algorithms, as well as assess their performance in order to select the most excellent for a given application. The advancement of the pattern recognition research community is measured on the performance obtained by the proposed pattern recognition systems on benchmark databases in fields such as biometrics, optical character recognition, medical images, object recognition, etc.

A challenging research topic in the field of Human-Machine Interaction is the analysis and recognition of emotional facial expressions. This is because recognizing faces (and in particular emotional faces) under gross environmental variations (such as the quality of the camera, light variations etcetera) and in real time remains a problem largely unsolved [3].

The collection and distribution of databases is a time-resource-consuming task, requiring experience and care both in the content design and the acquisition protocol. After the data collection, additional efforts are typically dedicated to supervise, annotate, label, error correct and document the collected data. In addition, a set of legal requirements have to be addressed, including consent forms to be signed by the donators and operational security measures as instructed by the data protection authorities. Finally, the distribution of the database involves intellectual property rights and maintenance issues.

When it comes to emotional facial expressions, according to the classical literature (largely debated but not yet superseded [7]) there are only 6 emotional categories[1] labelled as happiness, sadness, anger, fear, surprise, and disgust. However, the limited number of classes does not simplify the collection of a database of emotional facial expressions, due to the intrinsic difficulty to dispose of natural and spontaneous emotional samples from a significant amount of people. In order to collect such data there are mainly three procedures:

a) Recordings of spontaneous manifestations of emotional feelings. Generally this can be done by collecting video-recordings of subjects in their everyday activity, such as shopping, meeting, etcetera. The main drawback in such scenarios is the lack of control and therefore, a high amount of variability in the data, as well as the presence of few strikingly clear instances of episodic emotions. Cowie et al. [4], after the analysis of the Belfast naturalistic database, containing highly emotional talk-show recordings, showed that clear-cut emotional episodes were unexpectedly rare in such scenarios.

b) Recordings of subjects asked to simulate a specific facial emotional expression. Generally they are professional actors. However, although skilled actors can be convincing, it could be argued that they are not really experiencing the portrayed emotion, but a stylized version of the natural one, and therefore, a different set of facial features may be needed for their description. Batliner et al. [2] demonstrated that vocal signs of emotionality used by an actor simulating a particular human-machine interaction were different from, and much simpler than, those produced by people genuinely engaged in it.

c) Recordings of induced emotional states: to this aim there exists various emotion induction techniques. Some include the listening to emotional musical expressions, the watching of pictures and movies with highly emotional contents, as well as the playing of specially designed games. The advantage achieved in such scenarios is a higher situational control and thus, a major reliability of the collected data and the associated measurements.

---

[1] Not all the authors agree on these 6 (see [11] as an example)

An excellent overview of the different existing databases for the automatic modelling of emotional states is reported in [4].

According to the acquisition procedure, emotional databases can be split into four modalities: audio (typical measurements over speech signals are prosody, voice quality, timing, etc.), photos and video-sequences (eye-brow, and lip movements), gestures (hand and body movement) and physiological measures (temperature, humidity, heart rate, skin conductance, etc.). Some databases are collected accounting of several modalities simultaneously.

There is a considerable amount of image and audio emotional databases and a testimonial presence of physiological ones. Physiological measures of emotional states mainly refer to heart rates, skin temperature variations and electro-dermal activity. Such measurements always require the involved subject to wear a sensor which, no matter how comfortable it may be, can affect the physiological measurement.

It is worth mentioning on this respect the works of Kataoka et al. [12], Shusterman et al. [19] as well as Aubergé et al. [1] and Kim et al. [13] who implemented a 24-hour wearable ring or a wristwatch-type sensor to measure natural skin temperature (SKT) variations due to emotional stimuli.

The first to hypothesize that emotional feelings or stress may change the distribution of face temperature was Fumishiro [10] who used a thermal imager (the resolution was higher than 0.01ºC) to show that under emotional feeling the radiance temperature of eyes, noses and brows can vary in the range +/-0.2ºC. However, to date, there are no systematic studies linking face temperature and emotions. This work aims to scientifically test such a relationship by collecting a database of thermal and visible facial emotional expressions. The collected data will allow to assess if such changes can be considered an emotional feature and whether different emotions can be discriminated by different temperature values of the face or of regions of it.

To collect such data, emotions were induced though a carefully assessed experimental set-up (describe below) and the acquired database consisted of appropriately synchronized thermal and visible facial expressions. A selection of what were considered the most significantly emotional faces was also made using a custom developed Matlab software program. All the data, including those selected as best representatives of a given facial emotional expression, are available to the scientific community as part of the COST Action 2102 (http://cost2102.cs.stir.ac.uk/) activities. In addition, the present paper reports experimental data ascertaining the effects of emotions on memory word recognition tasks by measuring the individual recognition performance.

**2. Database design**

The aim of this work was to define a database of emotional facial expressions which could be considered as "much spontaneous as possible". The original idea was to identify video stimuli that could be used to elicit emotional states. Four emotions were selected among the six listed by Ekman [6] as basic emotions: *fear*, *happiness*, *sadness*, and *disgust*. A *neutral* state was also considered[2], intended here as a state where no emotion is induced. This was done for the practical reason to separate series of facial video sequences recorded under a given induced emotion from another one as well as, to control the effects of an emotional stimulus on the other. *Surprise* and *anger* were not considered, due to the difficulty to elicit such emotional states through video stimuli. The definition of the spontaneous emotional facial expression database passed through three steps: 1) The identification of video stimuli to elicit the emotions under consideration – i.e. how the video-clips were selected and assessed; 2) The identification of a memory word recognition task, acting as a distractive task for restoring the subject's neutral state; 3) The acquisition protocol.

### 2.1 Identification of video stimuli with high emotional content

A total of 60 video-clips[3], 10 for each of the abovementioned emotional states were downloaded from YouTube (www.youtube.it) using the emotion labels as keyword. The original audio-track was kept. These stimuli were assessed by 20 naïve Italian subjects (9 males and 11 females) asked to watch the video-clips (randomly presented through a PPT Presentation) and label them by using the most appropriate of the 5 abovementioned emotional categories or any other emotional label. In addition, subjects were asked to rate the intensity of the portrayed emotion by using a Likert scale [14] varying from 1 (very weak) to 5 (very strong) through the intermediate values of 2 (weak), 3 (medium), and 4 (quite strong).

The result of this assessment was the identification of 5 video-clips for each emotion category (happy, sad, disgust, fear), plus 5 short neutral video-clips (30 sec.) separating an emotional video from another in the same emotional category, and 3 long neutral (2 minutes) video-clips separating sequences of different emotional category. This amounted to a total of 28 selected video-clips constrained to an average intensity rate value no lower than 3.

### 2.2 Identification of the Memory Word Recognition Task

In order to avoid overlaps among the induced emotional categories, a word memory recognition task was defined. In literature [15] such tasks are also called "recognition" tasks and consist of: a) A learning phase, where the subject memorizes a list of words (in our case 8 Italian words); b) A retention phase, where the subject is involved in an activity that has nothing to do with the task (in our case she/he was watching a sequence of 5 emotional video-clips belonging to the same emotional category inter-lived with short neutral stimuli); c) A re-enactment phase in which the subject is presented with a new list of words and she/he must provide a YES (if the word was already in the word list previously seen) or NOT (otherwise) answer .

---

[2] The authors of the present paper do have reservations on the existence of a *neutral* natural feeling but the discussion is out of the scope of the present work.
[3] The selected video-clip length varied from 30 to 140 sec. Longer stimuli were needed to induce sadness and/or to restore the neutral feeling in the subjects.

To this aim 8 word lists were created, 4 named Memory Lists (ML) and 4 named Recognition Lists (RL) each containing 8 Italian words. Both the word lists were shown on a computer screen. In each RLi there were 4 words already presented in the associated MLi , i:=1, .., 4. Before the induction of any of the 4 abovementioned emotional states, the subject was asked to read and memorizes the words in an MLi. Then, she/he was asked to watch a sequence of 5 emotional video-clips all belonging to the same emotional category. Finally, the RLi associated to the previously presented MLi was presented to the subject and she/he was asked to indicate on a paper grid, whether or not the words in the RLi list were already in the previously seen MLi one. A total of 48 almost equally frequent bi-syllabic, and three-syllabic Italian words were selected from the *Corpus e Lessico di Frequenza dell'Italiano Scritto (CoLFIS)*, www.alphalinguistica.sns.it. CoLFIS contains about 3.798.275 Italian words classified for frequency and complexity (monosyllabic, bi-syllabic, three-syllabic, etcetera). The words were randomly assigned to the 8 lists, taking care that each RLi contained 4 words already included in the corresponding MLi i=1, …,4, the corresponding RLi i=1, …4. As an example the Ml1 and RL1 lists are reported in Table 1. The shared words are in bold.

**Table 1.** The ML1 and RL1 word lists. The shared words are in bold.

| ML1 ITALIAN WORDS | ML1 TRANSLATION | RL1 ITALIAN WORDS | RL1 TRANSLATION |
|---|---|---|---|
| LUOGO | **PLACE** | **CAMPO** | **FIELD** |
| **TITOLO** | TITLE | PIANO | FLOOR |
| VALORE | **VALUE** | METRO | METRE |
| **AZIONE** | ACTION | **LUOGO** | **PLACE** |
| **FUTURO** | FUTURE | SERIE | SERIES |
| PASSO | **STEP** | **PASSO** | **STEP** |
| CAMPO | **FIELD** | **VALORE** | **VALUE** |
| **SEGNO** | SIGN | PEZZO | PIECE |

### 2.3 Acquisition procedure: The experimental set up

The subject was invited to sit in front of a computer screen in order to perform the task which consisted of the following steps:
1. Read and memorize an MLi list in 30s;
2. Watch a set of 5 video-clips belonging to a given emotional category, each interleaved by a short neutral stimulus (N);
3. Read the RLi list associated to the previously seen MLi list;
4. Using a pencil and a YES or NOT answer signs the words in the RLi list seen in the MLi list;
5. Watch a Long Neutral (LN) stimulus;

6. Go back to step 2 until the end of the stimuli.

The stimuli presentation was randomized among the subjects according to the 4 different condition schemes reported in Table 2, where the letters indicate emotional categories, with S=sad, H=happy, F=fear, D=disgust. The facial expressions recorded from each subject were taken at 1 sec. sampling rate.

**Table 2.** Stimuli sequencing in each of the 4 identified CONDITIONS (A, B, C, and D). The letters indicate the emotional categories, with S=sad, H=happy, F=fear, D=disgust, N=Neutral. The number after the letter identifies the stimulus inside the category. For example, F3 indicates the third stimulus used for Fear. Note that the Neutral stimuli were always the same, but were associated randomly to the categories.

| | |
|---|---|
| **CONDITION A** | **ML1** F1N1F2N2F3N3F4N4F5N5 **RL1LN1 ML2** H1N1H2N2H3N3H4 N4 H5N5**RL2LN2 ML3**S1N1S2N2S3N3S4N4S5N5 **RL3 LN3 ML4** D1N1D2 N2D3N3D4N4D5N5 **RL4** |
| **CONDITION B** | **ML2** D1N1D2N2D3N3D4N4D5N5 **RL2LN2ML3** F1N1F2N2F3N3F4N4F5 N5**RL3LN3 ML4** H1N1H2N2H3N3H4N4H5N5 **RL4LN1ML1** S1N1S2 N2 S3N3S4N4S5N5 **RL1** |
| **CONDITION C** | **ML3** S1N1S2N2S3N3S4N4S5N5 **RL3LN3ML4** D1N1D2N2D3N3D4N4 D5N5 **RL4LN1ML1** F1N1F2N2F3N3F4N4F5N5 **RL1LN2 ML2** H1N1H2 N2H3N3H4N4H5N5 **RL2** |
| **CONDITION D** | **ML4** H1N1H2N2H3N3H4N4H5N5 **RL4LN1ML1**S1N1S2N2S3N3S4N4 S5N5**RL1LN2 ML2** D1N1D2N2D3N3D4N4D5N5**RL2LN3ML3** F1N1F2 N2F3N3F4N4F5N5 **RL3** |

**2.4 Hardware and software configuration**

The acquisition system was equipped with two cameras each one connected to a separate laptop. A thermal camera TESTO 880-3 was connected to laptop SONY VGN-NS21Z with CPU Intel Core 2 Duo P8600 2.4GHz, 4GB RAM, MS Vista. This camera provided a  160 × 120 pixel resolution and the temperature range was set to 23-38°C (this camera provides a NETD < 0.1ºC. NETD is the Noise Equivalent Temperature Difference, which is a measurement of the sensitivity of a detector of thermal radiation). The visible camera was a Logitech Webcam C250 connected to laptop SAMSUNG NP R519 LED/T4200/2GB/250GB/SHARED/15.4/ VHP. This camera was set for a 640 × 480 pixel image resolution. Yawcam software, version 0.3.3, was used on both laptops, for the acquisition.
In both cases an image file (with a timestamp in the name)  was acquired every 1 sec. in *.png format, in order to avoid huge memory space occupancy and consequently video compression. This is relevant especially for the thermal camera, sincethermal imagers do not provide as much resolution as webcams. Thus, it is important to keep the highest possible level of details without introducing a compression algorithm. The 1 sec. sampling rate was considered a good compromise between storage requirements and temporal-spatial resolution, since typical changes of muscular activities lasts for a few seconds [8].

**2.5 Acquisition scenario and timing**

The data collection was made in a quiet laboratory. Neither the acquisition computers, or the operators, or other people were visible to the participants. She/he watched the stimuli on a third laptop while wearing headphones to listen to the original video-clip audio-tracks, seated on a comfortable chair, with a black background and fluorescent room illumination, as illustrated in Figure 1. The acquisition took place from the 15$^{th}$ to the 19$^{th}$ of March 2010, between 9 a.m. and 6 p.m All the donors were Italian psychology undergraduate students, aged from 21 to 28 years. Such a population was deliberately chosen in order to reduce age and cultural background variability.

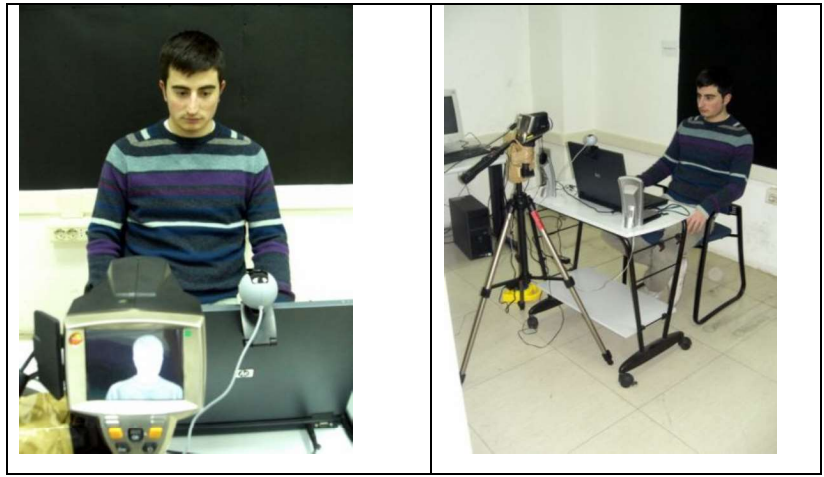

**Fig 1.** Acquisition scenario.

A consent form was filled and signed by each participant allowing the use of the collected data for scientific scopes. The acquisition timing for each subject is reported in Table 3

Table 3. Acquisition timing for each participant.

| Time elapsed | Tasks | Recording | Dialogue |
|---|---|---|---|
| 5 minutes | Explanation of "what to do". Signature of the consent form | No | Yes |
| 34s | Instructions on the computer screen | Yes | No |
| 37.85 minutes | Data collection according to Table 2 recording schema | Yes | No |
| 3 minutes | Subject comments and impressions | Yes | Yes |

## 2.6 Database description

More than 120.000 images for each camera (both the visual and thermal one) were collected during the experiment. Quantitative results obtained by human inspection are beyond the aims of this paper and may be tackled in future works.

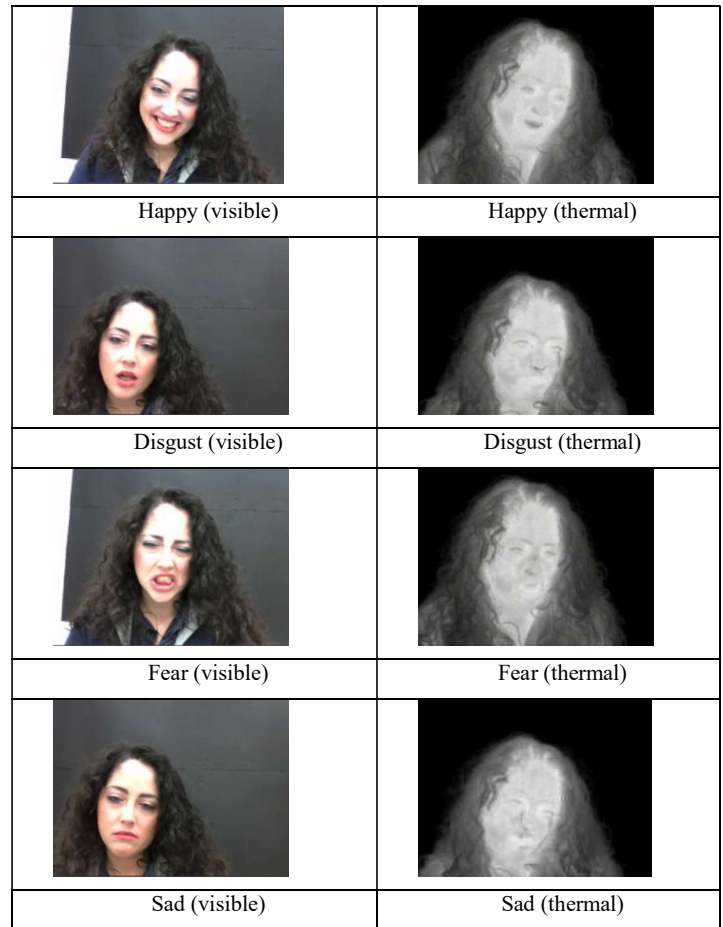

| Happy (visible) | Happy (thermal) |
| Disgust (visible) | Disgust (thermal) |
| Fear (visible) | Fear (thermal) |
| Sad (visible) | Sad (thermal) |

**Fig. 2** Visible and thermal examples of emotional facial expressions.

A snapshot in the visible and thermal domain of each induced emotional facial expression is displayed in Figure 2. It is worth noting that in the sad state the subject is crying and tears can clearly be seen in the thermal but not in the visible image.

## 2.7 Summary of the main characteristics

The collected database was named Italian Visible-Thermal Emotion (I.Vi.T.E.) database and will be freely distributed to the scientific community after the publication of this work. The main characteristics of the database are the following:
- Temperature range: 23-38 ºC;
- Size of database: 29.8GB (Thermal: 1.2GB, Visible: 28.6GB). Consists of one image per second in .png format;
- Total number of subjects: 49 Italian undergraduate students ranging from 22 to 28 years) watching sequences of highly emotional video-clips and listening to the original audio-tracks through headphones;
- Emotional categories under examination were: happiness, sadness, disgust, fear, and neutral;
- Thermal image resolution: 160 × 120 pixels;
- Visible image resolution: 640 × 480 pixels.
- Acquisition cameras: thermal (testo 880-3) and webcam (Logitech).

Using a custom Matlab software program the authors selected a total of 479 thermal and 479 visual images as the most significant facial emotional expressions elicited in the subjects. An example is displayed in Figure 3.

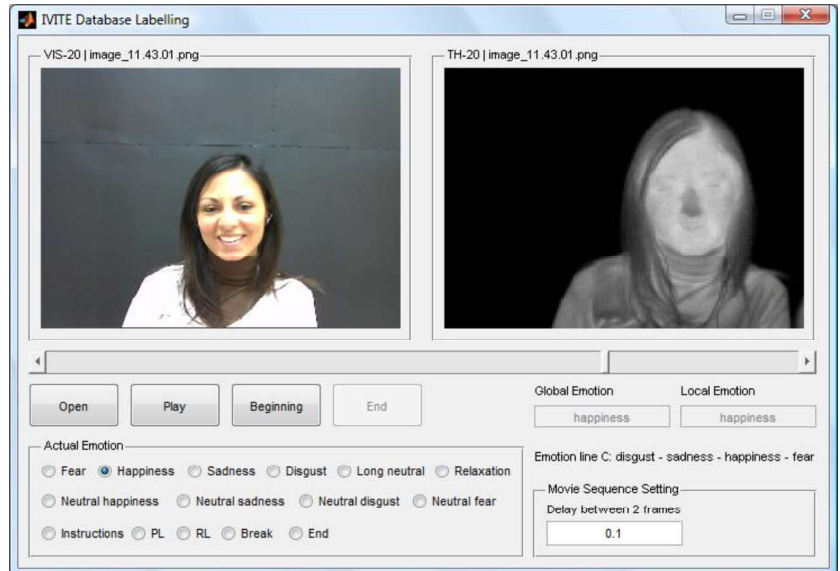

**Fig. 3.** A display of the software for manual labelling.

This work was necessary in order to eliminate, amidst all the captured images, those which, according to a couple of expert judges, did not belong to the emotional categories selected for the experiment. The software is able to name each selected image, showing the type of camera used (thermal or normal), the number assigned to the participant (1 to 49), the timestamp of the collected image (expressed in minutes, sec., and msec.) and the temperature in Celsius degrees reported by the thermal camera.

## 3. Results on the word memory recognition task

The effects of the emotional states on the word memory recognition task were assessed considering the averaged error committed by each subject on the RL lists, after watching a given sequence of emotional stimuli, all belonging to the same emotional category.

**Table 4.** The number of words in the RLi lists wrongly listed by the subjects in each experimental condition

|  | Condition A | Condition B | Condition C | Condition D |
|---|---|---|---|---|
| Fear | 18 | 16 | 5 | 12 |
| Disgust | 25 | 20 | 7 | 15 |
| Happiness | 13 | 18 | 24 | 14 |
| Sadness | 19 | 13 | 12 | 20 |

The original scores are reported in Table 4 for each of the four experimental conditions and for each emotion category. The numbers indicate how many words were wrongly listed in the RLi, i:=1,…4, lists by the subjects involved in a given experimental condition (A, B, C, D). Table 5 reports their transformation into z-scores with standard deviation $\sigma$ equal to 5,47.

**Table 5.** Z-score transformation of the data reported in Table 4 with $\sigma = 5,47$

| Z score | Condition A | Condition B | Condition C | Condition D |
|---|---|---|---|---|
| Fear | 0,42 | 0,06 | -1,95 | -0,67 |
| Disgust | 1,70 | 0,79 | -1,59 | -0,12 |
| Happiness | -0,49 | 0,42 | 1,52 | -0,30 |
| Sadness | 0,60 | -0,49 | -0,67 | 0,78 |

As exposed in Tables 5 and 6, there were no significant differences in the subject's memory performance that could be attributed to a given induced emotional state or to a given experimental condition. None of the Z-scores falls outside the average score distribution in the real interval of [-2, +2]. Even the C condition, where both the best and worst subjects' performance were gathered, does not show any significant deviation. The average word error in the word memory recognition task performed by the subjects on each emotional video sequence and for each of the 4 random elicited conditions is graphically displayed in Figure 4 and it varies in the real interval [0 2]. The average total error is illustrated in Figure 5. The data suggests that none of the induced emotional categories affects the word memory performance.

## 4. Results on the thermal data

The selected highly emotional faces, as reported in section 2.7, were manually tagged on 5 face regions (left part of left eye (LL), right part of left eye (RL), left part of right eye (LR), right part of right eye (RR), tip of nose (TN)) in order to measure possible changes in their temperature (with respect to the neutral state) when a given emotional state was induced.

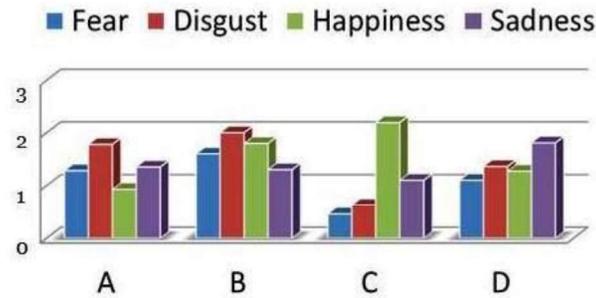

**Fig. 2.** Average word errors on the word memory recognition task distributed along the 4 experimental conditions.

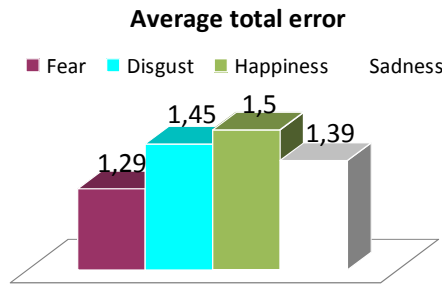

**Fig. 5.** Average total word errors on the word memory recognition task

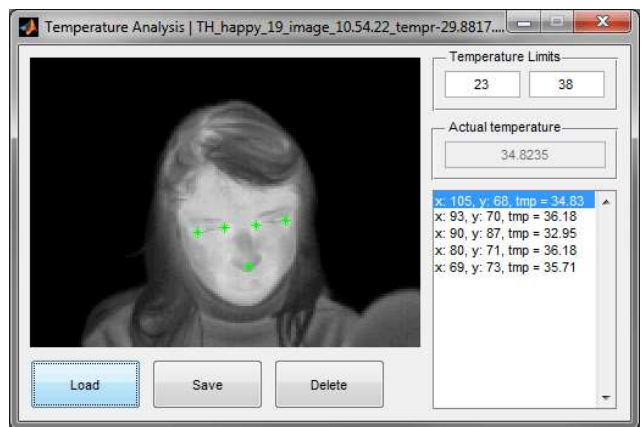

**Fig. 6.** Temperature analysis tool used for tagging the 5 face regions abovementioned

The custom Matlab software used for the tagging is illustrated in Figure 6. The temperature of these face regions was extracted from a 5x5 pixel matrix created around the selected points (as illustrated in Figure 6).

**Table 6.** Temperature data for the experimental condition A (females).

|  | Disgust | | | | | | Fear | | | | | |
|---|---|---|---|---|---|---|---|---|---|---|---|---|
| P. ID | WF | LL | RL | TN | LR | RR | WF | LL | RL | TN | LR | RR |
| 40 | -0.72 | -0.1 | -0.34 | -1.74 | -0.16 | -0.74 | 0.4 | 0.4 | 0.39 | 3.59 | 0.57 | 0.12 |
| 41 | -0.56 | -0.48 | -0.08 | -0.29 | -0.66 | 0.29 | 0.28 | -0.65 | 0.99 | 0.49 | 0.16 | 0.55 |
| 42 | -0.28 | 0.28 | 0.21 | -2.37 | -0.04 | -0.32 | 0.07 | -1.08 | -0.39 | 2.4 | -0.51 | -0.29 |
| 43 | -1.14 | 0.1 | -0.56 | -5.52 | 0.24 | -0.71 | 0.48 | -0.24 | -0.23 | 6.43 | -0.33 | 0.24 |
| 44 | 1.12 | -1.18 | 0.46 | 1.01 | 0.73 | -0.48 | -0.33 | -0.51 | -0.2 | -0.14 | -0.35 | 0.12 |
| mean | -0.32 | -0.28 | -0.06 | -1.78 | 0.02 | -0.39 | 0.18 | -0.41 | 0.11 | 2.56 | -0.09 | 0.15 |
| std | 0.86 | 0.58 | 0.41 | 2.47 | 0.51 | 0.42 | 0.32 | 0.55 | 0.57 | 2.63 | 0.45 | 0.3 |
|  | Happiness | | | | | | Sadness | | | | | |
| P. ID | WF | LL | RL | TN | LR | RR | WF | LL | RL | TN | LR | RR |
| 40 | -0.61 | 0.4 | 0.23 | -0.11 | 0.41 | -0.66 | -0.62 | -0.19 | -0.43 | -2.72 | -0.98 | -0.31 |
| 41 | -0.05 | 0.03 | -0.16 | -0.29 | -0.49 | 0.46 | -0.5 | 0.12 | 0.49 | 1.27 | 0.08 | 0.29 |
| 42 | -0.42 | -1.66 | 0.18 | 1.48 | 0.15 | -0.21 | 0.15 | 0.28 | 0.91 | 0.78 | 0.72 | -0.21 |
| 43 | 2.35 | 0.05 | 0.04 | 5.67 | 0.71 | 1.11 | 0.04 | 0.69 | 0.75 | 2 | 0.24 | -0.97 |
| 44 | -1.11 | -1.18 | 0.46 | -0.37 | -0.26 | -0.48 | -0.2 | -0.68 | -0.03 | -0.71 | 0.81 | -0.65 |
| mean | 0.03 | -0.47 | 0.15 | 1.28 | 0.1 | 0.05 | -0.23 | 0.05 | 0.34 | 0.12 | 0.17 | -0.37 |
| std | 1.35 | 0.9 | 0.23 | 2.57 | 0.49 | 0.73 | 0.33 | 0.51 | 0.56 | 1.88 | 0.72 | 0.47 |

**Table 7.** Temperature data for the experimental condition A (males).

|  | Disgust | | | | | | Fear | | | | | |
|---|---|---|---|---|---|---|---|---|---|---|---|---|
| P. ID | WF | LL | RL | TN | LR | RR | WF | LL | RL | TN | LR | RR |
| 46 | 2.28 | -0.03 | -0.36 | -3.52 | -0.22 | -0.57 | 1.14 | -0.37 | 1.45 | 0.56 | 0.66 | -1.14 |
| 47 | 1.44 | 0.46 | 0.57 | 0.3 | 0.66 | 0.53 | 0.35 | 0.96 | 0.32 | -0.88 | 0.74 | 1.13 |
| 48 | -0.81 | -1.45 | -0.91 | -6.07 | -0.78 | -2.81 | 0.09 | 0.14 | 0.77 | 7.08 | 0.69 | 0.86 |
| mean | 0.97 | -0.34 | -0.23 | -3.1 | -0.11 | -0.95 | 0.53 | 0.24 | 0.85 | 2.26 | 0.7 | 0.28 |
| std | 1.6 | 0.99 | 0.74 | 3.21 | 0.73 | 1.71 | 0.55 | 0.67 | 0.57 | 4.24 | 0.04 | 1.24 |
|  | Happiness | | | | | | Sadness | | | | | |
| P. ID | WF | LL | RL | TN | LR | RR | WF | LL | RL | TN | LR | RR |
| 46 | 0.13 | -0.03 | 0.79 | -1.8 | 0.66 | -0.97 | 0.18 | -0.46 | -0.53 | -2.94 | -0.58 | -0.54 |
| 47 | -0.56 | -0.2 | 0.32 | 0.3 | 0.58 | -0.07 | 1.06 | -0.53 | -0.48 | -1.21 | -0.55 | -1.27 |
| 48 | 0.62 | -0.4 | 0.6 | 3.69 | 0.69 | -0.57 | -1.67 | -1.4 | -0.46 | -4.02 | -0.46 | -1.25 |
| mean | 0.07 | -0.21 | 0.57 | 0.73 | 0.64 | -0.54 | -0.14 | -0.8 | -0.49 | -2.73 | -0.53 | -1.02 |
| std | 0.59 | 0.18 | 0.23 | 2.77 | 0.06 | 0.45 | 1.39 | 0.53 | 0.03 | 1.42 | 0.06 | 0.42 |

In addition, also the mean temperature of the whole face (WF) was considered. The measurements of the relative temperature changes (measured in Celsius degree relative changes with respect to the neutral state) are reported, as an exemplification and only for the experimental condition A, in Tables 6 and 7 for the females and males respectively. The gray columns indicate that 80% of the participants exhibited in such face regions a temperature change with respect to the neutral state, measured before the emotion was induced. However, these changes randomly appear in different face regions when the experimental conditions change from A to B, C, D.

**Table 8.** Temperature data for the experimental condition B (females).

|  | Disgust | | | | | | Fear | | | | | |
|---|---|---|---|---|---|---|---|---|---|---|---|---|
| P. ID | WF | LL | RL | TN | LR | RR | WF | LL | RL | TN | LR | RR |
| 6 | -0.42 | 1.04 | 0.22 | -5.85 | -0.44 | -0.54 | 0.24 | 1.19 | 0.59 | 2.26 | -0.49 | -0.59 |
| 7 | -0.8 | 1.79 | -0.2 | 1.5 | 0.08 | 0.46 | -1.37 | 1.14 | 0.56 | 2.21 | 0.65 | -0.75 |
| 8 | -0.86 | -1.43 | -0.81 | -3.48 | -1.08 | -1.28 | 0.02 | 1.21 | 0.32 | 6.44 | 0.22 | -0.11 |
| 11 | -0.03 | 0.17 | 0.08 | -0.93 | -0.17 | 0.21 | 0.01 | -0.68 | -0.76 | 1.13 | -0.59 | -0.98 |
| 12 | 0.61 | 0.23 | 0.03 | -1.59 | -0.26 | -0.63 | -0.6 | -0.92 | -0.53 | -0.74 | -0.26 | -0.63 |
| 14 | -0.76 | -0.13 | -0.3 | -1.7 | -0.47 | -0.37 | -0.17 | -0.05 | 0.2 | -0.4 | 0.51 | 0.97 |
| 15 | -0.02 | 0.24 | -0.02 | -4.12 | -0.21 | -0.23 | 0.13 | 0.09 | 0.37 | -1.21 | -0.18 | -0.04 |
| 16 | 0.49 | 0.9 | 0.15 | -1.29 | 0.06 | -0.55 | -1.02 | -1.1 | 0.23 | -1.18 | -0.02 | -0.55 |
| 17 | -1.44 | -0.77 | -0.03 | -1.23 | -0.46 | -0.7 | -0.58 | -0.29 | -0.08 | -0.91 | 0.11 | -0.21 |
| mean | -0.36 | 0.23 | -0.1 | -2.08 | -0.33 | -0.4 | -0.37 | 0.07 | 0.1 | 0.84 | -0.01 | -0.32 |
| std | 0.68 | 0.96 | 0.31 | 2.13 | 0.35 | 0.51 | 0.56 | 0.92 | 0.47 | 2.52 | 0.42 | 0.58 |
|  | Happiness | | | | | | Sadness | | | | | |
| P. ID | WF | LL | RL | TN | LR | RR | WF | LL | RL | TN | LR | RR |
| 6 | -0.02 | 0.22 | 0.11 | 5.98 | 0.06 | 0.11 | -0.36 | 0.8 | 0.3 | -4.32 | -0.06 | -0.28 |
| 7 | -0.71 | 1.06 | 0.08 | -1.46 | 0.73 | -1.2 | 1 | 0.32 | -1.13 | -1.81 | -1.01 | -1.12 |
| 8 | -0.53 | 1.21 | 0.24 | 5.11 | 0.14 | 0.14 | -1.63 | -0.19 | 0.08 | -4.19 | -0.11 | -1.36 |
| 11 | -0.21 | -0.59 | -0.42 | 2.37 | -0.34 | -0.21 | -0.11 | -0.68 | -0.76 | -0.51 | 0.25 | 1.58 |
| 12 | -1.07 | -0.51 | -0.05 | 0.02 | -0.34 | -0.71 | -0.07 | 0.23 | 0.11 | -1.34 | -0.26 | -0.13 |
| 14 | -0.31 | 0.03 | 0.03 | 0.73 | 0.59 | 0.64 | -0.71 | -0.91 | -0.24 | -1.52 | 0.04 | -0.37 |
| 15 | 0.15 | 0.85 | 0.86 | 0.12 | 0.85 | 0.62 | -0.98 | -0.41 | -0.12 | -1.94 | -0.18 | -0.1 |
| 16 | -0.17 | 0.34 | 0.39 | 0.79 | 0.82 | 1.77 | -0.23 | -0.6 | -0.37 | -0.72 | -0.15 | -0.08 |
| 17 | -1.01 | -1.07 | 0.25 | -0.69 | -0.77 | -0.71 | 1.35 | 0.88 | 0.36 | 2.17 | -0.05 | -0.6 |
| mean | -0.43 | 0.17 | 0.17 | 1.44 | 0.19 | 0.05 | -0.19 | -0.06 | -0.2 | -1.58 | -0.17 | -0.27 |
| std | 0.43 | 0.79 | 0.35 | 2.56 | 0.59 | 0.9 | 0.92 | 0.65 | 0.49 | 1.95 | 0.35 | 0.83 |

**Table 8.** Temperature data for the experimental condition B (males).

|  | Disgust | | | | | | Fear | | | | | |
|---|---|---|---|---|---|---|---|---|---|---|---|---|
| P. ID | WF | LL | RL | TN | LR | RR | WF | LL | RL | TN | LR | RR |
| 9 | -0.69 | -0.58 | 0.4 | -1.68 | 0.33 | 0 | -1.3 | 0.58 | 0.64 | 1.76 | 0 | -0.67 |
| 10 | 0.44 | -0.38 | -0.08 | -1.66 | 0.32 | 0.16 | 0.08 | -1.13 | -1.14 | -3.14 | -0.65 | 0 |
| 13 | -0.5 | -0.63 | -0.56 | -1.99 | -0.64 | -0.76 | -0.38 | -0.85 | 0.03 | -0.85 | 0 | 0.23 |
| mean | -0.25 | -0.53 | -0.08 | -1.77 | 0 | -0.2 | -0.53 | -0.47 | -0.15 | -0.74 | -0.22 | -0.15 |
| std | 0.61 | 0.13 | 0.48 | 0.19 | 0.56 | 0.5 | 0.7 | 0.92 | 0.91 | 2.45 | 0.37 | 0.47 |
|  | Happiness | | | | | | Sadness | | | | | |
| P. ID | WF | LL | RL | TN | LR | RR | WF | LL | RL | TN | LR | RR |
| 9 | -0.74 | -0.58 | 0.7 | 1.94 | -0.11 | -1.68 | 0.22 | -0.58 | -0.16 | -0.88 | -0.24 | 0.67 |
| 10 | 0.2 | 1.36 | 0 | -1.74 | 0.49 | 0.16 | -0.43 | -1.05 | -0.16 | -3.01 | -0.53 | 0.04 |
| 13 | 0.48 | 0.3 | 0.03 | -2.39 | -0.16 | 0.23 | -0.61 | 0.3 | 0.03 | -1.03 | 0 | 0.4 |
| mean | -0.02 | 0.36 | 0.24 | -0.73 | 0.07 | -0.43 | -0.27 | -0.44 | -0.1 | -1.64 | -0.26 | 0.37 |
| std | 0.64 | 0.97 | 0.39 | 2.34 | 0.36 | 1.09 | 0.43 | 0.68 | 0.11 | 1.19 | 0.26 | 0.32 |

For example, the temperature changes for the same emotional category in the experimental condition B (see Tables 8 and 9 for females and male respectively) do not follow the same pattern observed for the experimental condition A (see Tables 6 and 7).

Therefore, it seems that with this temperature resolution and in the defined experimental conditions, emotional states do not significantly change the temperature of the face or regions of it. An increased temporal-spatial resolution of the thermal camera to identify appreciable temperature changes would be necessary.

## 5.  Conclusions

This paper reports on a collection of naturalistic thermal and visible induced facial emotional expressions providing details on the experimental set-up, the acquisition scenario, the eliciting stimuli and the data. Facial emotional expression recognition through visible images has occupied a great deal of research, while thermal images have not yet been considered. Given that thermal images have the good property of not being affected by illumination and shadows, they can be, to a certain extend, more useful than the visible ones to determine distinctive facial emotional features.

In addition, this work reports data obtained through a pilot experiment, showing no effects of emotional states for a defined word memory recognition task. It could be argued that the proposed word memory recognition paradigm (memory task) proved to be ineffective by the emotional interference, compared to the recall paradigm proposed by Dougherty and Rauch [5]. However, this rises several open questions on the intervention of emotional states on memory performance. Further investigations are needed to assess which are, and to what extent cognitive and memory tasks are affected by emotional states. Some questions. Which emotional state will produce an improvement or a deterioration of the cognitive and memory performance? Does the feeling experienced by the subject in the learning or the retention phase play a role in the accuracy of the recognition? For a better memory performance, is the emotional feeling state at the time of the encoding more important than the one experienced during the retention of the mnemonic material? Literature suggests the importance of both [16-18]. However, more data are needed. Finally, what are the effects of the sequencing? Does it produce a bias in the learning and retention phase?

Finally, it was shown that an increased temporal-spatial resolution of the thermal camera would be necessary to observe appreciable temperature changes in the face or regions of it.